\documentclass[sigconf]{acmart}

\usepackage{booktabs} % For formal tables
\usepackage{bm}
\usepackage{color}
\usepackage{subfigure}
\usepackage{multirow}

\def\model{NHFM }

\def\E{\bm{E}} 
\def\e{\bm{e}}

\def\s{\bm{s}}

\def\x{\bm{x}}
\def\v{\bm{v}}

\def\setR{\mathbb{R}}
\def\model{NHFM}

\copyrightyear{2020} 
\acmYear{2020} 
\setcopyright{acmcopyright}\acmConference[SIGIR '20]{Proceedings of the 43rd International ACM SIGIR Conference on Research and Development in Information Retrieval}{July 25--30, 2020}{Virtual Event, China}
\acmBooktitle{Proceedings of the 43rd International ACM SIGIR Conference on Research and Development in Information Retrieval (SIGIR '20), July 25--30, 2020, Virtual Event, China}
\acmPrice{15.00}
\acmDOI{10.1145/3397271.3401307}
\acmISBN{978-1-4503-8016-4/20/07}

\begin{document}
\fancyhead{}
\title{Neural Hierarchical Factorization Machines \\ for User's Event Sequence Analysis}

\author{
Dongbo Xi$^{1,2,3}$,
Fuzhen Zhuang$^{1,2,*}$,
Bowen Song$^{3,*}$,
Yongchun Zhu$^{1,2,3}$,
Shuai Chen$^{3}$,
Dan Hong$^{3}$,
Tao Chen$^{3}$,
Xi Gu$^{3}$ and
Qing He$^{1,2}$
}
\affiliation{\institution{$^1$Key Lab of Intelligent Information Processing of Chinese Academy of Sciences (CAS),\\
Institute of Computing Technology, CAS, Beijing 100190, China}}
\affiliation{\institution{$^2$University of Chinese Academy of Sciences, Beijing 100049, China}}
\affiliation{\institution{$^3$Alipay (Hangzhou) Information \& Technology Co., Ltd.}}
\email{{xidongbo17s, zhuangfuzhen, zhuyongchun18s, heqing}@ict.ac.cn, {bowen.sbw, shuai.cs, dan.hong, boshan.ct, guxi.gx}@antfin.com}

\thanks{* Corresponding authors: Fuzhen Zhuang and Bowen Song.}

\def\authors{Dongbo Xi, Fuzhen Zhuang, Bowen Song, Yongchun Zhu, Shuai Chen, Dan Hong, Tao Chen, Xi Gu and Qing He}

\renewcommand{\shortauthors}{Xi and Zhuang, et al.}

\begin{abstract}
Many prediction tasks of real-world applications need to model multi-order feature interactions in user's event sequence for better detection performance. 
However,  existing popular solutions usually suffer two key issues: 
1) only focusing on feature interactions and failing to capture the sequence influence; 
2) only focusing on sequence information, but ignoring internal feature relations of each event, thus failing to extract a better event representation.
In this paper, we consider a two-level structure for capturing the hierarchical information over user's event sequence:
1) learning effective feature interactions based event representation;
2) modeling the sequence representation of user's historical events. 
Experimental results on both industrial and public datasets clearly demonstrate that our model achieves significantly better performance compared with state-of-the-art baselines.
\end{abstract}




\maketitle
\vspace{-1mm}
\section{Introduction}
The success of many prediction tasks (e.g., CTR prediction \cite{widedeep,he2017nfm,guo2017deepfm,lian2018xdeepfm,zhou2019dien}, fraud detection \cite{zhu2020modeling}, etc) in real-world applications to a large extent depend on mining effective features via analyzing user's rich historical event sequence information. 

Usually, in these prediction tasks, 
the user's historical behavior sequence is a list of events. 
In each event, rich categorical and numerical features (e.g., item ID, timestamp, etc) which describe the current event are available.
Therefore, how to efficiently take advantage of these internal features  inside each event and the sequence behavior information of each user is of fundamental importance for the prediction performance.

% Recently, considerable efforts have been made to address these tasks via mining rich raw features.
% one way is to learn feature interactions from the raw data to generate efficient feature representation.
Recently, a lot of efforts \cite{he2017nfm,guo2017deepfm,xiao2017afm,lian2018xdeepfm} have been done to capture multi-order feature interactions via combining the Factorization Machine (FM) \cite{rendle2010factorization} and neural network.
However, these efforts failed to capture user's  event sequence information. 
The other methods have also attempted to take the user's historical event sequence into consideration \cite{beutel2018lcrnn,zhou2019dien,tang2019m3r}.
Nevertheless, most of these existing studies mainly focused on the event sequence information but ignored the internal feature relations of each event.
Therefore, a long-standing challenge is how to effectively combine the event and sequence influence simultaneously for capturing hierarchical structure over user's event sequence.

Along this line, we propose a novel \textbf{Neural} \textbf{Hierarchical} \textbf{Factorization} \textbf{Machine} (\textbf{NHFM}) model.
For the first level, in each event, we apply an event extractor, which captures internal feature interactions via Hadamard product based FM, to extract an effective event representation.
For the second level, in the event sequence, we apply FM to capture event interactions.
Combining the two levels, we get the basic hierarchical interaction model NHFM-$\alpha$. 
% The NHFM-$\alpha$ does not add additional parameters compared with the NFM \cite{he2017nfm}, but it can model hierarchical structure for better performance.
Besides, for capturing user's historical event sequence and long range dependent information, we design an event self-importance-aware sequence module instead of the FM in the second level to form the NHFM-$\beta$.
The NHFM-$\beta$ takes the order and importance of different events into consideration.
Finally, we combine the NHFM-$\alpha$ and NHFM-$\beta$ in the second level to capture event interactions and self-importance sequence information simultaneously. 
\vspace{-1mm}
\section{Related Work}
\vspace{-1mm}
In this section, we present the related work in two-fold: feature interactions and user's event sequence analysis.

Factorization Machine (FM) \cite{rendle2010factorization} is a widely used method to model second-order feature interactions automatically.
Recently, some studies have also combined the advantages of the FM on modeling second-order and the neural network on modeling higher-order feature interactions \cite{he2017nfm,guo2017deepfm}. 
% For example, a method called Neural Factorization Machine (NFM) \cite{he2017nfm} has been proposed to use an FM followed by MLP.
However, these above efforts addressed the input as a single sparse vector and failed to capture user's event sequence information.

For most prediction tasks, it is necessary to capture user's event sequence information. 
A method called Multi-temporal-range Mixture Model (M3)
\cite{tang2019m3r} has been proposed to apply MLP to concatenated feature embedding vectors for extracting the event representation, and then employ a mixture of models to deal with both short-term and long-term dependencies.
Some other studies have also attempted to take user's historical event sequence into consideration \cite{beutel2018lcrnn,zhou2019dien,tang2019m3r}.
Nevertheless, these studies paid more attention to the event sequence information but ignored the internal feature relations of each event (e.g., they usually only utilize simple concatenation or MLP for the internal features of each event), which failed to obtain an effective event representation.
\vspace{-1mm}
\section{Methodology}
\vspace{-1mm}
\begin{figure}[!t]
	\begin{center}
		\includegraphics[width=0.8\linewidth]{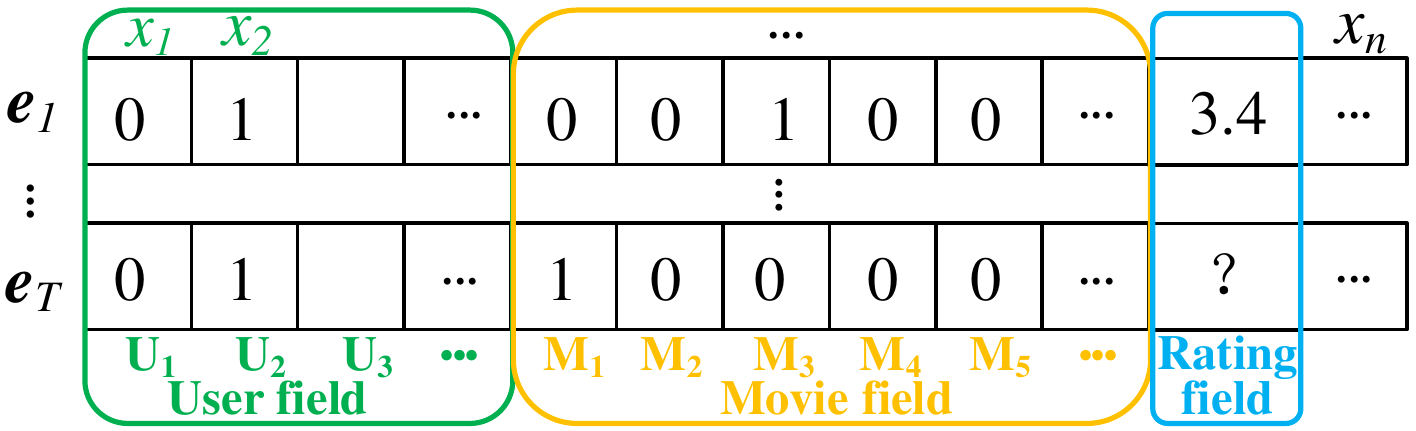}
		\vspace{-2mm}
		\caption{A simple example of event sequence $\E$.}
		\label{sample_example}
	\end{center}\vspace{-4mm}
\end{figure}
\begin{figure}[!t]
\begin{center}
\includegraphics[width=0.8\linewidth]{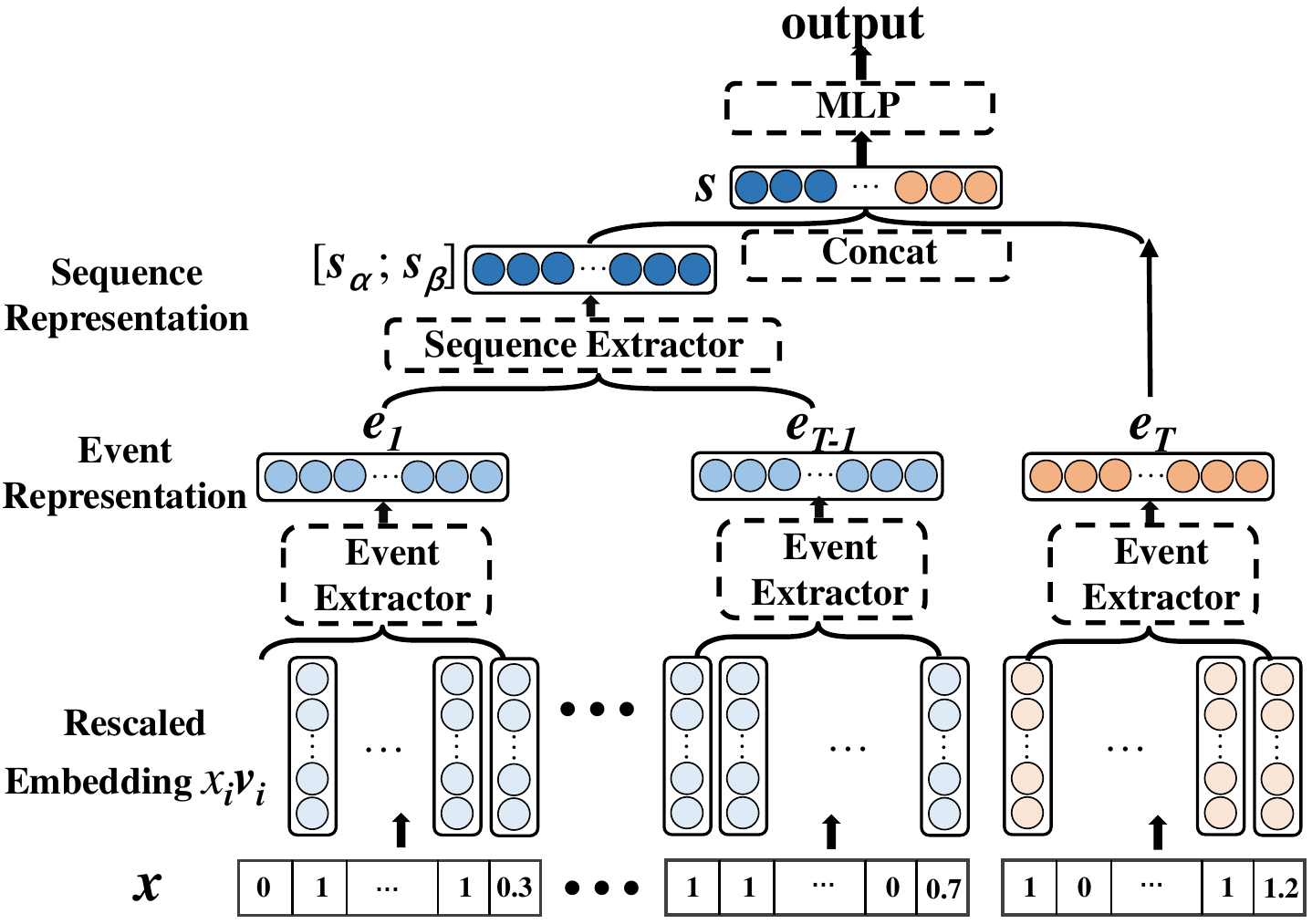}
\vspace{-2mm}
\caption{The proposed \model~model, for simplicity, we do not represent the ``wide" part.}
\label{fig:model}%
\end{center}\vspace{-4mm}
\end{figure}

In this section, we first formulate the problem, then present the details of the proposed \model~as shown in Figure \ref{fig:model}.
\vspace{-1mm}
\subsection{Problem Statement}
\vspace{-1mm}
Given a user's event sequence $\E = [\e_1, \e_2,..., \e_T]$, where $T$ is the event sequence length.
$\e_t = [x_1^t, x_2^t,..., x_n^t]$ $(1\leq t\leq T)$ is the $t$-th event of the user,
where $n$ is the size of feature dictionary. For categorical fields, $x_i^t$ $(1\leq i\leq n)$ is $1$ if $\e_t$ has the value in the current categorical field, otherwise is $0$. For numerical fields, $x_i^t$ is the normalized real value.
A simple example of event sequence $\E$ is shown in Figure \ref{sample_example}.
%These different fields describe each event.
The task is to make a prediction for the current event $\e_T$ according to the user's historical event sequence $[\e_1, \e_2,..., \e_{T-1}]$ and available information of the current event $\e_T$.
% and the task can be formulated as Classification, Regression or Pairwise Ranking via utilizing different activation and loss functions according to real application scenarios as described in \cite{rendle2010factorization}.
\vspace{-1mm}
\subsection{Event Extractor}
\vspace{-1mm}
As shown in Figure \ref{sample_example}, most of the features are one-hot encoding categorical features, the dimension is usually high and the vectors are sparse.
FM is an effective method to address such high-dimension and sparse problems.

Firstly, we project each non-zero feature $x_i$ to a low dimension dense vector $\v_i$.
The embedding layer learns one embedding vector $\v_i\in\setR^{k}$ ($1\leq i\leq n$) for each feature $x_i$,
where $n$ is the size of feature dictionary and $k$ is the dimension of embedding vectors. 
For both categorical and numerical features, we rescale the embedding via $x_i\v_i$. Therefore, we only need to include the non-zero features, i.e., $x_i\neq 0$.

Next, in order to extract internal feature interactions of each event for the event representation, we apply the FM for $t$-th event $\e_t$ as follow:
\vspace{-3mm}
\begin{eqnarray}
\e_t=FM(\x^t) = \sum_{i=1}^{n-1}\sum_{j=i+1}^n x^t_i\v_i\odot x^t_j\v_j.\label{equ:fm}
\end{eqnarray}
Different from traditional FM, which uses inner product to get a scalar,
we apply Hadamard product $\odot$ to get the vector representation of each event.
Hadamard product denotes the element-wise product of two vectors $(\v_i\odot\v_j)_k=v_{ik}v_{jk}$.

Now, we apply the event extractor (i.e., Equation (\ref{equ:fm})) to each event in the user's event sequence, and we get the event representations $\E = [\e_1, \e_2,..., \e_T]$ thereafter for the sequence extractor.

\vspace{-1mm}
\subsection{Sequence Extractor}
\vspace{-1mm}
The sequence extractor, which contains a parameter-free module and a module with parameters, can capture event interactions and event self-importance sequence information, respectively.

\subsubsection{Event Interactions Module}
In user behavior sequence analysis, user's final behavior is strongly correlated with the user's past several behaviors.
For example, in card-stolen fraud detection scenario, 
the fraudster's behaviors are associated with his/her abnomally frequent payment attempts and the changing of payment environment. 
Therefore, in this level, we adopt the FM to capture event interactions, combining with the feature interactions in the first level forming a hierarchical structure:
\begin{eqnarray}
\s_{\alpha}=FM(\E_{his}) = \sum_{i=1}^{T-2}\sum_{j=i+1}^{T-1} q_i\e_i\odot q_j\e_j,\label{equ:event_fm}
\end{eqnarray}
where $\E_{his}=[\e_1,...,\e_{T-1}]$ is the historical events,
$q_i\in\{0,1\}$ indicates whether the $i$-th event $\e_i$ exists.
The learned sequence representation $\s_{\alpha}$ is concatenated with the current prediction event $\e_T$ and then fed to an MLP to form our basic NHFM-$\alpha$ model.
 
It is worth pointing out that the NHFM-$\alpha$ does not introduce extra parameters\footnote{The total parameters of the model are in the embedding layer and MLP.} compared with the NFM \cite{he2017nfm}, and more importantly, it not only inherits the advantages of NFM on modeling second-order and higher-order feature interactions, but also could model the hierarchical structure information, which significantly improves its prediction performance comparing with the NFM.

\subsubsection{Event Self-importance-aware Sequence Module}
Even though the NHFM-$\alpha$ considers event interactions, it does not consider the event sequence information.
To address this problem, we design an event self-importance-aware sequence module instead of the FM in the second level of the NHFM-$\alpha$ to form the NHFM-$\beta$.
The module takes the relative importance and order information of different events into consideration.
Different historical events of users have different importance.
% For example, in card-stolen fraud detection scenario, 
% intuitively, user's payment event is more important than the sign in event.
Therefore, we design an event self-importance-aware attention to learn the importance of different events. 
% The key advantage of the event self-importance-aware attention is that it can capture the importance of each event in the event sequence.
The self-importance weight is defined as scaled dot-product:
\begin{eqnarray}
\hat{a}_t=\frac{<F_1(\e_t),F_2(\e_t)>}{\sqrt{k}},~~~~
a_t=\frac{\exp(\hat{a}_t)}{\sum_{t=1}^{T-1}\exp(\hat{a}_t)}, \label{equ:event_weight}
\end{eqnarray}
and the importance weighted event sequence is represented as:
$\s_{self}=\sum_{t=1}^{T-1}a_tF_3(\e_t)$,
where $F_1$, $F_2$ and $F_3$ represent the feed-forward
networks to project the input event vector to one new vector representation.

% Besides, Long Short-Term Memory (LSTM) \cite{hochreiter1997long} is capable of learning short and long-term dependencies and has become an effective and scalable model for sequential prediction.
Besides, for learning short and long-term dependencies, we adopt the bidirectional LSTM (Bi-LSTM) \cite{graves2013speech} whose basic idea is to present each sequence forward and backward to the recurrent net, both of which are summed to get the sequence representation:
$
\s_{RNN}=Bi-LSTM(\E_{his}),
$
where $\s_{RNN}\in\setR^h$ and $h$ is the dimension of the LSTM output.
For every event in the event sequence, the network has complete sequential information before and after it, so the network can capture long range event dependent information.
The output of the event self-importance-aware sequence module is the concatenation of $\s_{self}$ and $\s_{RNN}$:
\begin{eqnarray}
\s_{\beta}=[\s_{self};\s_{RNN}].
\end{eqnarray}

Finally, we combine the NHFM-$\alpha$ and NHFM-$\beta$ in the second level to capture event interactions and self-importance sequence information simultaneously, then they are concatenated with current prediction event $\e_T$ and fed to an MLP. 
The output of the MLP is combined with a linear ``wide'' part, and then fed to the activation function to form the final NHFM model:
\begin{eqnarray}
\s=[\s_{\alpha};\s_{\beta};\e_T],~~~~
\hat{y}=sigmoid(MLP(\s)+f(\x)),
\end{eqnarray}
where $[\s_{\alpha};\s_{\beta}]$ models user's history behaviors (e.g., in fraud detection, it can be regarded as the risk of user's historical behaviors, and in recommender systems, it can be regarded as user's historical preference), and $\e_T$ is user's current risk or preference representation.
Besides, $f(\x)$ is the ``wide'' part just like the part in Wide\&Deep \cite{widedeep}:
$
f(\x)=\sum_{t=1}^T\sum_{i=1}^nw_ix_i^t+w_0, 
$
where $w_i$ indicates the importance of the feature $x_i$.
For binary classification tasks, we need to minimize the negative log-likelihood.
% \begin{equation}
% L(\theta)=-\frac{1}{N}\sum^N_{(\x,y)\in\D}(\left(y\log\hat{y}+(1-y)\log(1-\hat{y})\right),
% \end{equation}
% where $N$ is the number of samples, $y$ is the label of sample $\x$ and $\theta$ is the parameter set.
\vspace{-2mm}
\section{Experiments}
\vspace{-1mm}
\begin{table}[!t]
  \centering
  \caption{Summary statistics for the datasets.}
  \vspace{-2mm}
%   \resizebox{0.75\linewidth}{0.12\linewidth}{
\setlength{\tabcolsep}{0.5mm}{
    \begin{tabular}{cccccc}
    \toprule
          & Dataset & \#pos & \#neg & \#fields & \#events \\
    \midrule
    \multirow{3}[2]{*}{Industrial} & C1    & 15K   & 1.37M & 56    & 4.28M \\
          & C2    & 10K   & 1.93M & 56    & 3.57M \\
          & C3    & 5.7K  & 174K  & 56    & 353K \\
    \midrule
    Public & Movielens-1M & 575K  & 425K  & 9    & 1M \\
    \bottomrule
    \end{tabular}
    }
  \label{tab:dataset}\vspace{-4mm}
\end{table}%
In this section, we perform experiments to evaluate the \model~ against state-of-the-art methods on both industrial and public datasets. 
\vspace{-1mm}
\subsection{Experimental Setup}
\vspace{-1mm}
\textbf{Datasets.}
The industrial datasets contain the card transaction samples from one international e-commerce platform.
We utilize three countries (\textbf{C1},\textbf{ C2}, \textbf{C3}) data.
The task is to detect whether the current payment event is a card-stolen case.
The public dataset is the MovieLens-1M\footnote{http://www.grouplens.org/datasets/movielens/} dataset.
We binarize the ratings following the common process \cite{pan2019warm}. 
We regard each user's rating for each movie as an event.
The statistics of all datasets are shown in Table \ref{tab:dataset}.

\textbf{Baselines.}
We compare the proposed method with feature interactions based models (\textbf{W\&D}~\cite{widedeep}, \textbf{DeepFM}~\cite{guo2017deepfm},  \textbf{NFM}~\cite{he2017nfm},  \textbf{AFM}~\cite{xiao2017afm},  \textbf{xDeepFM}~\cite{lian2018xdeepfm}) and event sequence based models (\textbf{LSTM4FD}~\cite{wang2017session},   \textbf{LCRNN}~\cite{beutel2018lcrnn},  \textbf{M3}~\cite{tang2019m3r}). 

\textbf{Evaluation Metrics.}
We use the standard metric: \textbf{AUC} (Area Under ROC).
Besides, in our real card-stolen fraud detection scenario, 
we should increase the recall rate of the fraudulent transactions, in the same time, disturbing as few normal users as possible.
In other words, we need to improve the True Positive Rate (TPR) on the basis of low False Positive Rate (FPR).
Therefore, we also adopt the standardized partial AUC (\textbf{AUC$\bm{_{FPR\leq1\%}}$}) \cite{mcclish1989auc} (The standardized area of the head of ROC curve when the FPR$\leq 1\%$).
For all experiments, we report
the above metrics with 95\% confidence intervals on five random runs.
``*'' indicates that the improvement is statistically significant compared with the best baselines at p-value < 0.05 over independent samples t-tests.

\begin{table}[!t]
  \centering
  \caption{AUC$_{FPR\leq1\%}$ and AUC performance (mean$\pm$95\% confidence intervals) on datasets of C1, C2, C3 and MovieLens.}
  \vspace{-2mm}
  \resizebox{0.92\linewidth}{!}{
    \begin{tabular}{c||c||c||c||c}
    \toprule
    \multirow{2}[2]{*}{Model} & C1    & C2    & C3    & MovieLens \\
          & AUC$_{FPR\leq1\%}$ & AUC$_{FPR\leq1\%}$ & AUC$_{FPR\leq1\%}$ & AUC \\
    \midrule
    W\&D  & 0.6997$\pm$0.0014 & 0.7770$\pm$0.0023 & 0.8202$\pm$0.0092 & 0.7553$\pm$0.0006 \\
    DeepFM & 0.7070$\pm$0.0060 & 0.7725$\pm$0.0035 & 0.8476$\pm$0.0070 & 0.7651$\pm$0.0023 \\
    NFM   & 0.7467$\pm$0.0032 & 0.7931$\pm$0.0074 & 0.8310$\pm$0.0221 & 0.7583$\pm$0.0007 \\
    AFM   & 0.7086$\pm$0.0047 & 0.8061$\pm$0.0065 & 0.8499$\pm$0.0073 & 0.7536$\pm$0.0028 \\
    xDeepFM & 0.7391$\pm$0.0044 & 0.7833$\pm$0.0084 & 0.8557$\pm$0.0137 & 0.7643$\pm$0.0021 \\
    \midrule
    LSTM4FD & 0.7118$\pm$0.0087 & 0.7356$\pm$0.0084 & 0.7762$\pm$0.0086 & 0.7595$\pm$0.0001 \\
    LCRNN & 0.7089$\pm$0.0068 & 0.7847$\pm$0.0109 & 0.8162$\pm$0.0191 & 0.7613$\pm$0.0009 \\
    M3   & 0.7294$\pm$0.0060 & 0.7897$\pm$0.0053 & 0.7618$\pm$0.0268 & 0.7638$\pm$0.0003 \\
    \midrule
    NHFM-$\alpha$ & 0.7643$\pm$0.0054 & 0.8268$\pm$0.0071 & 0.8524$\pm$0.0090 & 0.7678$\pm$0.0005 \\
    NHFM-$\beta$ & 0.7697$\pm$0.0048 & 0.8369$\pm$0.0046 & 0.8628$\pm$0.0101 & 0.7696$\pm$0.0003 \\
    NHFM  & \textbf{0.7753$\pm$0.0037*} & \textbf{0.8551$\pm$0.0055*} & \textbf{0.8742$\pm$0.0089*} & \textbf{0.7708$\pm$0.0006*} \\
    \bottomrule
    \end{tabular}
    }
  \label{tab:result1}\vspace{-4mm}
\end{table}%
\vspace{-1mm}
\subsection{Performance Comparison}
\vspace{-1mm}
The experimental results are presented in Table \ref{tab:result1}. 
From these results, we have the following insightful observations:

- For the feature interactions based models, the performance of W\&D is inferior compared with other baselines, perhaps because it can not automatically learn feature interactions. 
Considering all datasets, the performance of other feature interactions based models is similar. They all attempt to capture high-order feature interactions via combining the FM and neural network but ignore the event sequence information. 

- For the event sequence based models, the overall performance of M3 and LCRNN is slightly better than LSTM4FD due to theirs more complex improved models based on the LSTM.

- For C3 dataset, it has fewer positive, negative samples and available events than C1 and C2 as shown in Table \ref{tab:dataset}.
Therefore, the AUC$_{FPR\leq1\%}$ performance of these models varies a lot, and the confidence intervals on C3 dataset is obviously larger than the C1 and C2 datasets.
For the public MovieLens dataset, the overall AUC performance of different baselines varies very little. It makes sense that the public dataset have fewer fields in each event as shown in Table \ref{tab:dataset}, so the feature interactions can not play his role well.
Therefore, the feature interactions based and event sequence based baseline models achieve almost similar performance.

- We also construct ablation experiments over NHFM-${\alpha}$, NHFM-${\beta}$ and \model.
NHFM-${\alpha}$ and NHFM-${\beta}$ utilize Event Interactions Module and Event Self-importance-aware Sequence Module in their sequence extractor, respectively.
The proposed \model~combines the NHFM-${\alpha}$ and NHFM-${\beta}$ in the sequence extractor and obtains the significantly best performance, which indicates both two modules matter for the performance.

These improvements also indicate that the proposed \model~can better handle the task via capturing the hierarchical structure.
\vspace{-2mm}
\subsection{Case Study}
\vspace{-1mm}
\begin{table}[!t]
  \centering
  \caption{The extracted high risk and low risk features, the ``Amount" is of payment, the ``Expiration" is of bank cards (Year), the ``Time" is  the time when the event occurs (Hour: Minute) and the ``Category" is of goods.}
  \vspace{-2mm}
\resizebox{0.85\linewidth}{!}{
    \begin{tabular}{ccccc}
    \toprule
          & Amount (\$)& Expiration& Time & Category  \\
    \midrule
          & 1500-2000 & 2035  & 03:00  &     \\
      High   & 1000-1500 & 2037  & 01:00  &  Digital  \\
      Risk   & 800-1000 & 2036  & 05:00  &     \\
         
    \midrule
     & 0-20 & 2022  & 17:00 &   \\
     Low      &20-40 & 2021  & 12:00 &  Common    \\
      Risk    &40-60 & 2025  & 14:00 &       \\ 
    \bottomrule
    \end{tabular}%
    }
  \label{tab:case1}\vspace{-3mm}
\end{table}%
In this subsection, we make some analysis on the interpretability of the proposed \model~model.

Firstly, in the feature level, we extract some high-risk and low-risk features according to the learned weights in the ``wide'' part.
We present the features in Table \ref{tab:case1}.
From these results, we have the following findings,
1) The larger the amount range, the higher the risk.
2) The expiration of bank cards is usually less than 10 years, so the cards whose expiration is more than 10 years are abnormal and high-risk.
3) Fraudulent transactions usually take place in the middle of the night, so the risk is higher at $00:00-06:00$.
4) In e-commerce platform, digital goods belong to high-risk category, so the fraud risk is higher than that of common goods.
\begin{figure}[!t]
	\begin{center}
		\includegraphics[width=1\linewidth]{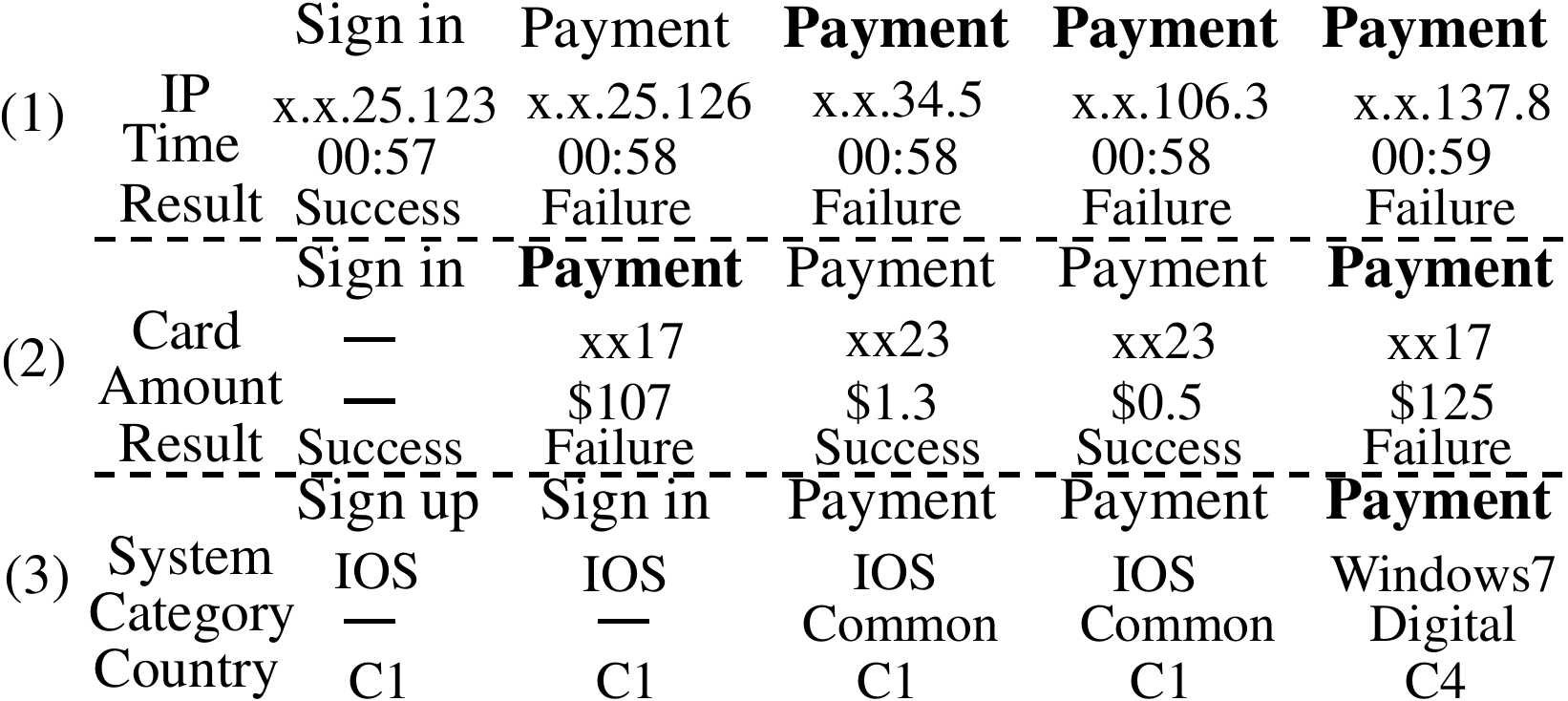}
		\vspace{-4mm}
		\caption{The extracted high-risk event sub-sequences from positive samples, the bold events have higher weights.}
		\label{fig:case2}
	\end{center}\vspace{-5mm}
\end{figure}
Then, in the event level, we extract some high-risk event sub-sequences from positive samples (the fraud label is 1) according to the learned event self-importance weights in Equation (\ref{equ:event_weight}).
These event sub-sequences can be regarded as the modeling of users' behavior patterns.
We present the behavior patterns in Figure \ref{fig:case2}.
First, for sub-sequence (1), after the first payment fails, the user modified the IP several times in a minute, which led to multiple payment failures and had a high risk. Therefore, the last three payment events obtain higher weights.
Then, for sub-sequence (2), after the failure of the first large payment, the user tried two small payments and succeeded with another card. Then he tried to make a large payment with the card he used for the first time, but failed.
Therefore, the two large payments obtain higher weights.
In sub-sequence (3), 
the account was initially used by a normal user and purchased common goods.
Then, the account was stolen by the user with different device in another country to buy high-risk digital goods, so the last payment event has a higher weight.

These results show that our \model~ can effectively find the important features and events for mining high-risk behavior patterns.
\vspace{-4mm}
\section{Conclusion}
\vspace{-1mm}
In this paper, we designed a \model~for event sequence analysis. 
Specially, we applied event and sequence extractors as a hierarchical structure
to learn event and sequence representations.
The experimental results on industrial and public datasets show significant improvement compared with various state-of-the-art baselines. 
\vspace{-1mm}
\begin{acks}
\vspace{-1mm}
The research work supported by the National Key Research and Development Program of China under Grant No. 2018YFB1004300, the National Natural Science Foundation of China under Grant No. U1836206, U1811461, 61773361, the Project of Youth Innovation Promotion Association CAS under Grant No. 2017146. This work is funded in part by Ant Financial through the Ant Financial Science Funds for Security Research. We also thank Minhui Wang, Zhiyao Chen, Changjiang Zhang for their valuable suggestions.
\end{acks}
\vspace{-1mm}
\bibliographystyle{ACM-Reference-Format}
\balance
\bibliography{sigir20_short}


\begin{thebibliography}{14}

%%% the Web address.

\ifx \showCODEN    \undefined \def \showCODEN     #1{\unskip}     \fi
\ifx \showDOI      \undefined \def \showDOI       #1{#1}\fi
\ifx \showISBNx    \undefined \def \showISBNx     #1{\unskip}     \fi
\ifx \showISBNxiii \undefined \def \showISBNxiii  #1{\unskip}     \fi
\ifx \showISSN     \undefined \def \showISSN      #1{\unskip}     \fi
\ifx \showLCCN     \undefined \def \showLCCN      #1{\unskip}     \fi
\ifx \shownote     \undefined \def \shownote      #1{#1}          \fi
\ifx \showarticletitle \undefined \def \showarticletitle #1{#1}   \fi
\ifx \showURL      \undefined \def \showURL       {\relax}        \fi
% invisible to TeX
\providecommand\bibfield[2]{#2}
\providecommand\bibinfo[2]{#2}
\providecommand\natexlab[1]{#1}
\providecommand\showeprint[2][]{arXiv:#2}

\bibitem[\protect\citeauthoryear{Beutel, Covington, Jain, Xu, Li, Gatto, and
  Chi}{Beutel et~al\mbox{.}}{2018}]%
        {beutel2018lcrnn}
\bibfield{author}{\bibinfo{person}{Alex Beutel}, \bibinfo{person}{Paul
  Covington}, \bibinfo{person}{Sagar Jain}, \bibinfo{person}{Can Xu},
  \bibinfo{person}{Jia Li}, \bibinfo{person}{Vince Gatto}, {and}
  \bibinfo{person}{Ed~H Chi}.} \bibinfo{year}{2018}\natexlab{}.
\newblock \showarticletitle{Latent cross: Making use of context in recurrent
  recommender systems}. In \bibinfo{booktitle}{\emph{WSDM}}.
\newblock


\bibitem[\protect\citeauthoryear{Cheng, Koc, Harmsen, Shaked, Chandra, Aradhye,
  Anderson, Corrado, Chai, Ispir, et~al\mbox{.}}{Cheng et~al\mbox{.}}{2016}]%
        {widedeep}
\bibfield{author}{\bibinfo{person}{Heng-Tze Cheng}, \bibinfo{person}{Levent
  Koc}, \bibinfo{person}{Jeremiah Harmsen}, \bibinfo{person}{Tal Shaked},
  \bibinfo{person}{Tushar Chandra}, \bibinfo{person}{Hrishi Aradhye},
  \bibinfo{person}{Glen Anderson}, \bibinfo{person}{Greg Corrado},
  \bibinfo{person}{Wei Chai}, \bibinfo{person}{Mustafa Ispir}, {et~al\mbox{.}}}
  \bibinfo{year}{2016}\natexlab{}.
\newblock \showarticletitle{Wide \& deep learning for recommender systems}. In
  \bibinfo{booktitle}{\emph{DLRS}}.
\newblock


\bibitem[\protect\citeauthoryear{Graves, Mohamed, and Hinton}{Graves
  et~al\mbox{.}}{2013}]%
        {graves2013speech}
\bibfield{author}{\bibinfo{person}{Alex Graves}, \bibinfo{person}{Abdel-rahman
  Mohamed}, {and} \bibinfo{person}{Geoffrey Hinton}.}
  \bibinfo{year}{2013}\natexlab{}.
\newblock \showarticletitle{Speech recognition with deep recurrent neural
  networks}. In \bibinfo{booktitle}{\emph{ICASSP}}.
\newblock


\bibitem[\protect\citeauthoryear{Guo, Tang, Ye, Li, and He}{Guo
  et~al\mbox{.}}{2017}]%
        {guo2017deepfm}
\bibfield{author}{\bibinfo{person}{Huifeng Guo}, \bibinfo{person}{Ruiming
  Tang}, \bibinfo{person}{Yunming Ye}, \bibinfo{person}{Zhenguo Li}, {and}
  \bibinfo{person}{Xiuqiang He}.} \bibinfo{year}{2017}\natexlab{}.
\newblock \showarticletitle{DeepFM: a factorization-machine based neural
  network for CTR prediction}. In \bibinfo{booktitle}{\emph{IJCAI}}.
\newblock


\bibitem[\protect\citeauthoryear{He and Chua}{He and Chua}{2017}]%
        {he2017nfm}
\bibfield{author}{\bibinfo{person}{Xiangnan He} {and} \bibinfo{person}{Tat-Seng
  Chua}.} \bibinfo{year}{2017}\natexlab{}.
\newblock \showarticletitle{Neural factorization machines for sparse predictive
  analytics}. In \bibinfo{booktitle}{\emph{SIGIR}}.
\newblock


\bibitem[\protect\citeauthoryear{Lian, Zhou, Zhang, Chen, Xie, and Sun}{Lian
  et~al\mbox{.}}{2018}]%
        {lian2018xdeepfm}
\bibfield{author}{\bibinfo{person}{Jianxun Lian}, \bibinfo{person}{Xiaohuan
  Zhou}, \bibinfo{person}{Fuzheng Zhang}, \bibinfo{person}{Zhongxia Chen},
  \bibinfo{person}{Xing Xie}, {and} \bibinfo{person}{Guangzhong Sun}.}
  \bibinfo{year}{2018}\natexlab{}.
\newblock \showarticletitle{xdeepfm: Combining explicit and implicit feature
  interactions for recommender systems}. In \bibinfo{booktitle}{\emph{KDD}}.
\newblock


\bibitem[\protect\citeauthoryear{McClish}{McClish}{1989}]%
        {mcclish1989auc}
\bibfield{author}{\bibinfo{person}{Donna~Katzman McClish}.}
  \bibinfo{year}{1989}\natexlab{}.
\newblock \showarticletitle{Analyzing a portion of the ROC curve}.
\newblock \bibinfo{journal}{\emph{Medical Decision Making}}
  (\bibinfo{year}{1989}).
\newblock


\bibitem[\protect\citeauthoryear{Pan, Li, Ao, Tang, and He}{Pan
  et~al\mbox{.}}{2019}]%
        {pan2019warm}
\bibfield{author}{\bibinfo{person}{Feiyang Pan}, \bibinfo{person}{Shuokai Li},
  \bibinfo{person}{Xiang Ao}, \bibinfo{person}{Pingzhong Tang}, {and}
  \bibinfo{person}{Qing He}.} \bibinfo{year}{2019}\natexlab{}.
\newblock \showarticletitle{Warm Up Cold-start Advertisements: Improving CTR
  Predictions via Learning to Learn ID Embeddings}. In
  \bibinfo{booktitle}{\emph{SIGIR}}.
\newblock


\bibitem[\protect\citeauthoryear{Rendle}{Rendle}{2010}]%
        {rendle2010factorization}
\bibfield{author}{\bibinfo{person}{Steffen Rendle}.}
  \bibinfo{year}{2010}\natexlab{}.
\newblock \showarticletitle{Factorization machines}. In
  \bibinfo{booktitle}{\emph{ICDM}}.
\newblock


\bibitem[\protect\citeauthoryear{Tang, Belletti, Jain, Chen, Beutel, Xu, and
  H~Chi}{Tang et~al\mbox{.}}{2019}]%
        {tang2019m3r}
\bibfield{author}{\bibinfo{person}{Jiaxi Tang}, \bibinfo{person}{Francois
  Belletti}, \bibinfo{person}{Sagar Jain}, \bibinfo{person}{Minmin Chen},
  \bibinfo{person}{Alex Beutel}, \bibinfo{person}{Can Xu}, {and}
  \bibinfo{person}{Ed H~Chi}.} \bibinfo{year}{2019}\natexlab{}.
\newblock \showarticletitle{Towards neural mixture recommender for long range
  dependent user sequences}. In \bibinfo{booktitle}{\emph{TheWebConf}}.
\newblock


\bibitem[\protect\citeauthoryear{Wang, Liu, Gao, Qu, and Xu}{Wang
  et~al\mbox{.}}{2017}]%
        {wang2017session}
\bibfield{author}{\bibinfo{person}{Shuhao Wang}, \bibinfo{person}{Cancheng
  Liu}, \bibinfo{person}{Xiang Gao}, \bibinfo{person}{Hongtao Qu}, {and}
  \bibinfo{person}{Wei Xu}.} \bibinfo{year}{2017}\natexlab{}.
\newblock \showarticletitle{Session-Based Fraud Detection in Online E-Commerce
  Transactions Using Recurrent Neural Networks}. In
  \bibinfo{booktitle}{\emph{PKDD}}.
\newblock


\bibitem[\protect\citeauthoryear{Xiao, Ye, He, Zhang, Wu, and Chua}{Xiao
  et~al\mbox{.}}{2017}]%
        {xiao2017afm}
\bibfield{author}{\bibinfo{person}{Jun Xiao}, \bibinfo{person}{Hao Ye},
  \bibinfo{person}{Xiangnan He}, \bibinfo{person}{Hanwang Zhang},
  \bibinfo{person}{Fei Wu}, {and} \bibinfo{person}{Tat-Seng Chua}.}
  \bibinfo{year}{2017}\natexlab{}.
\newblock \showarticletitle{Attentional factorization machines: Learning the
  weight of feature interactions via attention networks}. In
  \bibinfo{booktitle}{\emph{IJCAI}}.
\newblock


\bibitem[\protect\citeauthoryear{Zhou, Mou, Fan, Pi, Bian, Zhou, Zhu, and
  Gai}{Zhou et~al\mbox{.}}{2019}]%
        {zhou2019dien}
\bibfield{author}{\bibinfo{person}{Guorui Zhou}, \bibinfo{person}{Na Mou},
  \bibinfo{person}{Ying Fan}, \bibinfo{person}{Qi Pi}, \bibinfo{person}{Weijie
  Bian}, \bibinfo{person}{Chang Zhou}, \bibinfo{person}{Xiaoqiang Zhu}, {and}
  \bibinfo{person}{Kun Gai}.} \bibinfo{year}{2019}\natexlab{}.
\newblock \showarticletitle{Deep interest evolution network for click-through
  rate prediction}. In \bibinfo{booktitle}{\emph{AAAI}}.
\newblock


\bibitem[\protect\citeauthoryear{Zhu, Xi, Song, Zhuang, Chen, Gu, and He}{Zhu
  et~al\mbox{.}}{2020}]%
        {zhu2020modeling}
\bibfield{author}{\bibinfo{person}{Yongchun Zhu}, \bibinfo{person}{Dongbo Xi},
  \bibinfo{person}{Bowen Song}, \bibinfo{person}{Fuzhen Zhuang},
  \bibinfo{person}{Shuai Chen}, \bibinfo{person}{Xi Gu}, {and}
  \bibinfo{person}{Qing He}.} \bibinfo{year}{2020}\natexlab{}.
\newblock \showarticletitle{Modeling Users’ Behavior Sequences with
  Hierarchical Explainable Network for Cross-domain Fraud Detection}. In
  \bibinfo{booktitle}{\emph{TheWebConf}}.
\newblock


\end{thebibliography}
\end{document}